\documentclass[10pt,twocolumn,letterpaper]{article}

\usepackage{cvpr}              

\usepackage[accsupp]{axessibility}  
\usepackage{graphicx}
\usepackage{amsmath}
\usepackage{amssymb}
\usepackage{booktabs}

\usepackage[utf8]{inputenc} 
\usepackage{url}            
\usepackage{booktabs}       
\usepackage{amsfonts}       
\usepackage{nicefrac}       
\usepackage{microtype}      
\usepackage{lipsum}
\usepackage{xcolor}
\usepackage{multirow}
\usepackage{array}
\usepackage{pifont}
\usepackage{rotating} 
\usepackage{amssymb}
\usepackage{amsmath}
\usepackage{tabularx}
\usepackage[switch]{lineno}
\graphicspath{ {./images/} }

\usepackage[pagebackref,breaklinks,colorlinks]{hyperref}

\usepackage[capitalize]{cleveref}
\crefname{section}{Sec.}{Secs.}
\Crefname{section}{Section}{Sections}
\Crefname{table}{Table}{Tables}
\crefname{table}{Tab.}{Tabs.}

\def\cvprPaperID{10979} 
\def\confName{CVPR}
\def\confYear{2022}

\begin{document}

\title{Vision-Language Pre-Training with Triple Contrastive Learning}

\author{
    Jinyu Yang\textsuperscript{\rm 1}, Jiali Duan\textsuperscript{\rm 2}, Son Tran\textsuperscript{\rm 2}, Yi Xu\textsuperscript{\rm 2}, Sampath Chanda\textsuperscript{\rm 2}, Liqun Chen\textsuperscript{\rm 2},
    Belinda Zeng\textsuperscript{\rm 2},\\ Trishul Chilimbi\textsuperscript{\rm 2}, 
    and Junzhou Huang\textsuperscript{\rm 1}
    \\
    \textsuperscript{\rm 1}University Of Texas at Arlington,
    \textsuperscript{\rm 2}Amazon\\
    {\tt\small jinyu.yang@mavs.uta.edu}, {\tt\small jzhuang@uta.edu}\\
    {\tt\small \{duajiali,sontran,yxaamzn,csampat,liquchen,zengb,trishulc\}@amazon.com}
}
\maketitle

\begin{abstract}
   Vision-language representation learning largely benefits from image-text alignment through contrastive losses (e.g., InfoNCE loss). The success of this alignment strategy is attributed to its capability in maximizing the mutual information (MI) between an image and its matched text. However, simply performing cross-modal alignment (CMA) ignores data potential within each modality, which may result in degraded representations. For instance, although CMA-based models are able to map image-text pairs close together in the embedding space, they fail to ensure that similar inputs from the same modality stay close by. This problem can get even worse when the pre-training data is noisy. In this paper, we propose triple contrastive learning (TCL) for vision-language pre-training by leveraging both cross-modal and intra-modal self-supervision. Besides CMA, TCL introduces an intra-modal contrastive objective to provide complementary benefits in representation learning. To take advantage of localized and structural information from image and text input, TCL further maximizes the average MI between local regions of image/text and their global summary. To the best of our knowledge, ours is the first work that takes into account local structure information for multi-modality representation learning. Experimental evaluations show that our approach is competitive and achieves the new state of the art on various common down-stream vision-language tasks such as image-text retrieval and visual question answering.
\end{abstract}


\section{Introduction}
Self-supervision is an active research topic both in vision and language representation learning. Numerous methods have been proposed with an impressive performance on challenging tasks \cite{he2020momentum,chen2020simple,chen2021empirical,devlin2018bert,radford2018improving,gao2021simcse}. \footnotetext[1]{\url{https://github.com/uta-smile/TCL}}\footnotetext[2]{This work was done while Jinyu Yang was interning at Amazon}
A typical approach is to pre-train a model on massive amounts of unlabeled data in a self-supervised manner, then fine-tune it for downstream tasks (e.g., zero-shot learning and transfer learning) of interest.
In the vision, self-supervision can be carried out using exemplars \cite{dosovitskiy2015discriminative}, predicting the relative position between two random patches \cite{doersch2015unsupervised} or via solving jigsaw \cite{noroozi2016unsupervised}.
In language, masked language modeling (MLM) is widely used as the method of choice for self-supervision.

\begin{figure*}[t]
  \centering
  \includegraphics[width=1\linewidth]{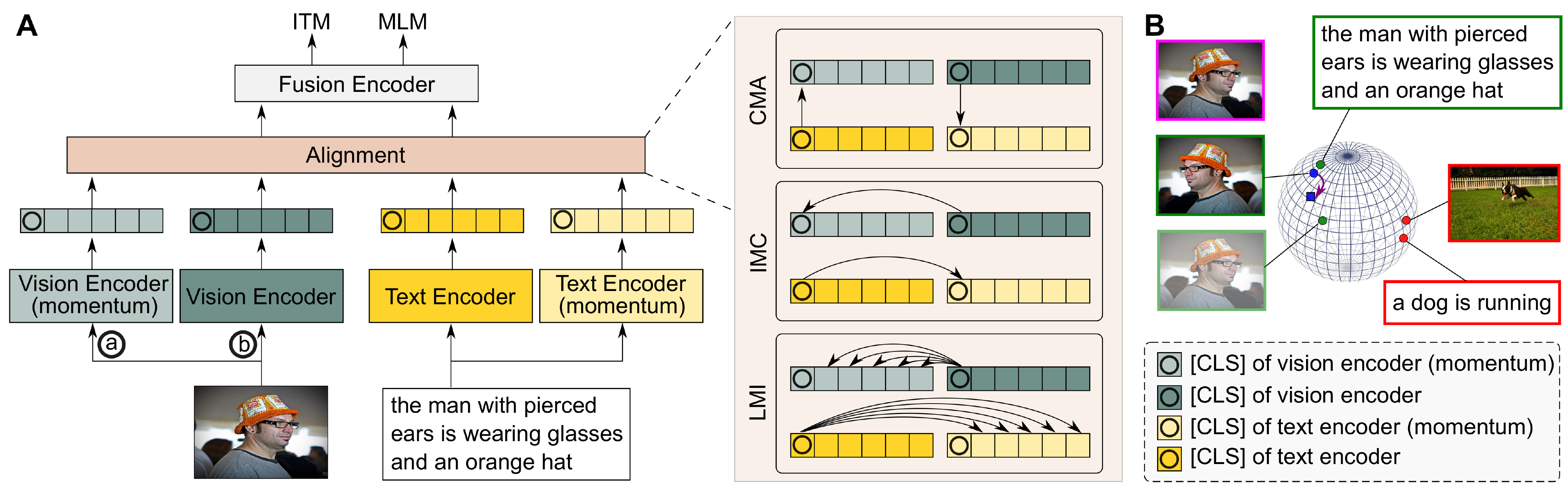}
   \caption{(A): An overview of our framework which consists of a vision encoder, a text encoder, and a fusion encoder. Each encoder has a paired momentum encoder updated by the momentum-based moving average.
   For the image input, we apply two separate data augmentation operators (a and b) which are sampled from the same family of augmentations. The alignment module contains three contrastive objectives (i.e., CMA, IMC, and LMI) for both cross-modal and intra-modal representation learning (make it easier for the fusion encoder to learn joint multi-modal embeddings). (B): The motivation of leveraging both cross-modal and intra-modal supervision. The original image (pink) is augmented to two different views (green). For CMA only, the middle image only has a positive text example (green) and treats other texts (red) as negatives. Its embedding (blue cirble) would be close to its positive text example. By incorporating IMC, it has two positive examples (one text and one image) and two sets of negative examples (one from text and one from image) and tends to learn more reasonable embeddings (blue square).
   } 
   \label{fig:framework}
   \vspace{-0.2in}
\end{figure*}

Inspired by the success of self-supervision in individual modalities, there is a surging interest in self-supervised vision-language pre-training (VLP) \cite{chen2022vlp,du2022survey}, which is essential for multi-modal tasks such as visual question answering (VQA), image-text retrieval, and visual entailment.
These tasks heavily rely on joint multi-modal embeddings which are typically obtained by modeling interactions between vision and language features.
To achieve this goal, various VLP frameworks are proposed by exploiting massive image-text pairs in the past few years \cite{li2020oscar,li2020unimo,gan2020large,chen2020uniter}, where the key insight is to apply a fusion encoder to the concatenation of vision and language features to learn joint representations.
Although simple and effective, this strategy suffers from the problem that vision and language features lie in different embedding spaces, which makes the feature fusion quite challenging \cite{li2021align}.
To mitigate this problem, the most recent state of the art \cite{li2021align} disentangles the learning process into two stages: i) first align cross-modal features by using a contrastive loss (i.e., InfoNCE \cite{oord2018representation}) to pull the embeddings of matched image-text pairs together while pushing those of non-matched pairs apart; then ii) apply a fusion encoder to the aligned image and text representations to learn joint embeddings.
Specifically, stage 1 aims to maximize the mutual information (MI) between matched image-text pair $(I, T)$ through InfoNCE loss, which is spurred by the fact that $I$ and $T$ represent two "views" of the same semantic \cite{wang2016learning}.
However, the limitation of stage 1 lies in that: simply performing cross-modal alignment (CMA) cannot fully guarantee the expressiveness of the learned features that is essential for joint multi-modal representation learning.  
The main reason is that $I$ and $T$ are unable to fully describe each other.
For instance, the text in (Figure~\ref{fig:framework} A) only focus on salient objects in the paired image, while overlooking other detailed and fine-grained information.
To align $I$ and $T$, only co-occurring features are captured by CMA.
This is also evidenced by \cite{jia2021scaling}, where the performance of CMA-based features on image-text retrieval is far greater than intra-modal retrieval (image-image and text-text).
Furthermore, the pre-training datasets are usually collected from the web and are inherently noisy. This leads to learning degraded representations, where cross-modal features fail to capture certain key concepts.

As transformer became increasingly popular in both vision and language tasks, existing VLP methods adopted transformer architecture for extracting visual and linguistic features.
Specifically, [CLS] tokens from the vision transformer (e.g., ViT \cite{dosovitskiy2020image}) and the text transformer (e.g., BERT \cite{devlin2018bert}) are used to represent the global information of the input.
For instance, ALBEF \cite{li2021align} maximizes MI between vision [CLS] and text [CLS].
However, global MI maximization fails to consider localized and structural information in the input \cite{hjelm2018learning,bachman2019learning}.
One potential side-effect is that it encourages the encoder to mainly extract information from certain unrelated/noisy image patches or text tokens that dominate MI.

In this paper, we introduce a novel VLP framework called triple contrastive learning (or TCL for short).
The key idea is to learn desirable representations by leveraging both cross-modal and intra-modal self-supervision, aiming to make it easier for the fusion encoder to learn multi-modal interactions.
To achieve this goal, TCL introduces three contrastive modules: cross-modal alignment (CMA), intra-modal contrastive (IMC), and local MI maximization (LMI), all of which rely on MI maximization. 
Specifically, i) CMA pulls the embeddings
of matched image-text pairs together while pushing those of non-matched pairs apart by maximizing global MI between matched image and text; ii) complementary to CMA, IMC maximizes agreement between differently augmented views of the same data example through maximizing their global MI; iii) LMI encourages high MI between the global representation and every local region (e.g., image patches and text tokens) of the input, which is designed to remedy the side-effects that are introduced by the global MI maximization.
The combination of these three modules allows us to i) learn representations that are semantically meaningful not only for cross-modal image-text pairs but also for intra-modal inputs; ii) capture the structural and localized information by extracting relevant features that are shared across local patches/tokens.


Our main contributions can be summarized as 
\begin{itemize}
\item We leverage both cross-modal and intra-modal self-supervision to provide complementary benefits in representation learning, which facilitates modeling of better joint multi-modal features in the fusion encoder;
\item Rather than simply relying on global information for multi-modal contrastive learning, we propose to take advantage of localized and structural information in both image and text input by maximizing local MI maximization between local regions and their global summary;
\end{itemize}

Comprehensive empirical studies demonstrate that TCL achieves a new state of the art on a wide range of vision+language benchmarks, such as image-text retrieval and VQA.
Specifically, on zero-shot image-text retrieval tasks, our method achieves significant improvement than ALIGN \cite{jia2021scaling}
(a mean recall of 79.5\% vs 70.9\% on MSCOCO). 
It is noteworthy that ALIGN is pre-trained on 1.8B image-text pairs, which is approximately 350$\times$ larger than TCL (5M).
By pre-training TCL on a large-scale dataset with 14M image-text pairs, we observe a significant performance boost, implying its potential for further improvement with larger datasets.
To investigate the effectiveness of each component in TCL, comprehensive ablation studies are also carried out with detailed analyses.

\section{Related Work}
\paragraph{Vision-Language Pre-training (VLP)}
Inspired by the success of self-supervised learning in intra-modal tasks, there is a surging interest in developing pre-training objectives for tasks with multiple modalities (e.g., vision and language).
For instance, to leverage a much broader source of supervision from text, pioneering work CLIP \cite{radford2021learning} predicts which text goes with which image, resulting in a task-agnostic model that is even competitive with task-specific supervised models.
ALIGN \cite{jia2021scaling} further scales up CLIP by leveraging a noisy dataset that covers more than one billion image alt-text pairs.
Despite these advancements, CLIP and ALIGN are mainly designed for vision-based downstream tasks and ignore the interaction between multiple modalities during pre-training.
To fit in vision+language tasks (such as VQA \cite{goyal2017making} and visual reasoning), recent studies propose to learn joint multi-modal representations of image content and natural language.
Among them, OSCAR \cite{li2020oscar}, UNIMO \cite{li2020unimo}, VILLA \cite{gan2020large}, and UNITER \cite{chen2020uniter} use an object detector (e.g., Faster R-CNN \cite{ren2015faster}) to capture vision features first, then a multi-layer transformer \cite{vaswani2017attention} is applied to the concatenation of the extracted vision features and text features to learn joint embeddings.
However, such kind of strategy suffers from limitations such as i) extracting region features using an object detector is computationally inefficient and ii) the quality of visual features is largely limited by the predefined visual vocabulary in pre-trained object detectors.
To address this issue, rather than rely on region-based visual features, SOHO \cite{huang2021seeing} takes a whole image as input and extracts compact image features through a visual dictionary, which favors 10 times faster inference time than region-based methods.
ViLT \cite{kim2021vilt} totally discards convolutional visual features and adopts vision transformer \cite{dosovitskiy2020image} to model long-range dependencies over a sequence of fixed-size non-overlapping image patches.
Although these aforementioned methods achieve remarkable performance, they fail to conduct image-text alignment before fusion, which makes it challenging to learn the interaction between different modalities.

To remedy this, ALBEF \cite{li2021align} applies a contrastive loss to align image and text features before modeling their joint representations, which delivers the state-of-the-art performance.
Our method shares similar spirits with ALBEF, but with clear differences as follows: 
i) instead of only performing cross-modal alignment (CMA), we propose to leverage both cross-modal and intra-modal self-supervision to enforce the learned representations are semantic meaningful.
The rationale is that cross-modal alignment alone may result in feature degeneration problem: although features from different modalities are well-separated, those from the same modality fall within a narrow cone and have high similarity.
ii) we introduce local alignment to the cross-modal scenario by maximizing mutual information (MI) between local regions and global representations.
Compared with the global alignment strategy used in ALBEF, maximizing local MI encourages our model to learn features that are shared across image patches/text tokens.
Furthermore, local alignment prevents simply capturing noise or unrelated features.

CODIS \cite{duan2022multi} is a concurrent work, which adopts a teacher-student distillation paradigm to guide the learning process.
Different from our method, CODIS performs feature alignment using cluster representations.

\vspace{-10pt}
\paragraph{Mutual Information Maximization}
Mutual information (MI) aims to measure the relationship between random variables or determine the amount of shared information.
MI is widely used in unsupervised feature learning, where the key idea is to maximize MI between the input and output \cite{linsker1988self}.
However, the MI estimation of high dimensional random variables is quite difficult and intractable \cite{paninski2003estimation}, especially for deep neural networks.
To this end, MINE \cite{belghazi2018mine} exploits dual optimization to offer a general-purpose estimator of MI.
Another alternative is InfoNCE \cite{oord2018representation}, which is a categorical cross-entropy loss that identifies the positive sample amongst a set of negative samples.
InfoNCE is proved to be a lower bound of MI \cite{oord2018representation}, such that minimizing InfoNCE loss can indirectly maximize MI.
However, existing studies simply maximize MI between the complete input and output (i.e., global MI maximization), which is proven to be insufficient for meaningful representation learning \cite{hjelm2018learning}.  
DIM \cite{hjelm2018learning} addresses this limitation by introducing local MI, i.e., maximizing the average MI between local regions of image input and the encoder output. 
AMDIM \cite{bachman2019learning} further extends DIM to maximize MI between features extracted from independent augmentations of the same image.

Both DIM and AMDIM are conducted in intra-modal tasks. In contrast, we introduce local MI maximization for multi-modal problems to benefit cross-modal representation learning.
Specifically, we encourage high
MI between the global representation and every local region
(e.g., image patches and text tokens) of the input.
This enables more transferable representations which are evidenced by our empirical studies.
Furthermore, rather than using a CNN-based network, our local MI is built upon the transformer architecture.
Therefore, the sequential patch tokens in the transformer actually give us free access to local features without the need to pull local information from intermediate layers.
Our experiments show that patch embeddings from the last layer outperform intermediate-layer patches with the transformer backbone.

\section{Method}
In this section, we first describe the model architecture of our method (Figure~\ref{fig:framework}), followed by the uni-modal representation learning.
After that, we detail the proposed triple contrastive learning modules: cross-modal alignment (CMA), Intra Modal Contrastive (IMC), and Local MI maximization (LMI).
In the end, we brief two pre-training objectives, i.e., image-text matching (ITM) and masked language modeling (MLM).

\subsection{Model Architecture}
An overview of our method is shown in Figure~\ref{fig:framework}, which contains a vision encoder $g(\cdot)$ for learning visual features from the image input, a text encoder $h(\cdot)$ for learning linguistic features from the text input, and a fusion encoder for learning multi-modal interactions.
All of these encoders adopt transformer-based architecture \cite{vaswani2017attention}, which are detailed in section \ref{implementation_details}.  
For each encoder, we maintain a paired momentum encoder that is implemented by momentum-based moving average strategy by following the same setting in \cite{he2020momentum}.
Formally, $\theta_{\hat{g}} = m \theta_{\hat{g}}+ (1-m)\theta_g$, where $\hat{g}(\cdot)$ is momentum vision encoder, $m \in [0,1]$ is a momentum coefficient. 
Similarly, we use $\hat{h}(\cdot)$ to denote the momentum text encoder.
Uni-modal encoders $g(\cdot)$ and $h(\cdot)$ are used to learn robust visual and linguistic features from the given input, then an alignment module is applied to the learned features to align both cross-modal and intra-modal representations before fusion.
We detail each component in the following sections.

\subsection{Uni-modal Representation Learning}
Given an image-text pair $(I, T)$, two separate augmentations are applied to the image to obtain two correlated “views”, i.e., $I_1$ and $I_2$. 
Following \cite{chen2020simple,he2020momentum}, we consider two random “views” of the same image under random data augmentation as a positive pair.
Each augmented image is split into fixed-size patches which are then linearly mapped and embedded with positional information \cite{dosovitskiy2020image}.
Similar to BERT \cite{devlin2018bert}, a class token [CLS] is prepended to the image patches, serving as the representation of the whole image. 
The obtained sequential embeddings of $I_1$ are finally fed into $g(\cdot)$ to learn desired visual representations $\{v_{cls}, v_1, ..., v_M\}$, where $M$ is the total number of image patches.
For $I_2$, we use $\hat{g}(\cdot)$ to learn its representations $\{\hat{v}_{cls}, \hat{v}_1, ..., \hat{v}_M\}$.
For the text input $T$, we follow \cite{devlin2018bert} to obtain $\{{t}_{cls}, {t}_1, ..., {t}_N\}$ and $\{\hat{t}_{cls}, \hat{t}_1, ..., \hat{t}_N\}$ by $h(T)$ and $\hat{h}(T_+)$, where $N$ is the length of text tokens, $T_+=T$.

To model the interaction between image and text features, previous VLP work directly apply a fusion encoder to the concatenation of $\{v_{cls}, v_1, ..., v_M\}$ and $\{{t}_{cls}, {t}_1, ..., {t}_N\}$ to learn joint multi-modal embeddings.
However, the most obvious drawback of this strategy is that visual and linguistic features lie in different embedding spaces, which is challenging for the fusion encoder to learn their interactions \cite{li2021align}.
To alleviate this limitation, we propose an alignment module that is applied to the learned visual and linguistic features before fusion.
Specifically, our alignment module contains three contrastive learning objectives, i.e., CMA, IMC, and LMI.
We discuss each objective below and show that they play a complementary role in the feature alignment and can benefit multi-modal feature fusion.


\subsection{Cross-Modal Alignment (CMA)}
The goal of CMA is to pull embeddings of the matched image-text pair (sampled from the joint distribution) together while pushing those of unmatched pairs apart (sampled from the product of marginal distributions).
In other words, CMA aims to maximize the MI between the image and text that are matched, which are assumed to describe the same semantic meaning.
For instance, the text in Figure~\ref{fig:framework} (A) describes high-level information (e.g., the occurrence of certain events or presence of certain objects) in the paired image.
Since direct maximization of MI for continuous and high-dimensional variables is intractable \cite{belghazi2018mine}, we instead minimize InfoNCE loss \cite{oord2018representation} which represents the lower bound of MI.
Formally, the InfoNCE loss for image-to-text is defined as:
\begin{equation} \label{eq:i2t}
    \begin{aligned}
    \mathcal{L}_{nce}(I_1, T_+, \tilde{T})= - \mathbb{E}_{p(I, T)} \bigg[ log \frac{e^{(\text{sim}(I_1, T_+)/\tau)}}{\sum_{k=1}^K e^{(\text{sim}(I_1, \tilde{T}_k)/\tau)}} \bigg]
    \end{aligned}
\end{equation}
where $\tau$ is a temperature hyper-parameter, $\tilde{T}=\{\tilde{T}_1, ..., \tilde{T}_K\}$ is a set of negative text examples that are not matched to $I_1$, $\text{sim}(I_1,T_+)=f_v(v_{cls})^T \hat{f}_t(\hat{t}_{cls})$, where $f_v(\cdot)$ and $\hat{f}_t(\cdot)$ are two projection heads that map representations to the space where InfoNCE loss is applied.
To maintain the negative text samples $\tilde{T}$, following \cite{li2021align}, we use a large queue that keeps the most recent $K$ projected representations $\hat{f}_t(\hat{t}_{cls})$.
Similarly, the loss of text-to-image is formulated by:
\begin{equation} \label{eq:t2i}
    \begin{aligned}
    \mathcal{L}_{nce}(T, I_2, \tilde{I})= - \mathbb{E}_{p(I, T)} \bigg[ log \frac{e^{(\text{sim}(T, I_2)/\tau)}}{\sum_{k=1}^K e^{(\text{sim}(T, \tilde{I}_k)/\tau)}} \bigg]
    \end{aligned}
\end{equation}
where $\text{sim}(T,I_2)=f_t(t_{cls})^T \hat{f}_v(\hat{v}_{cls})$, $f_t(\cdot)$ and $\hat{f}_v(\cdot)$ are two projection heads.
$\tilde{I}=\{\tilde{I}_1, ..., \tilde{I}_K\}$ is a queue of negative image examples which store the most recent $K$ projected features $\hat{f}_v(\hat{v}_{cls})$.
Taken together, we define the loss of CMA as:
\begin{equation}
    \begin{aligned}
    \mathcal{L}_{cma}=\frac{1}{2}[\mathcal{L}_{nce}(I_1, T_+, \tilde{T})+\mathcal{L}_{nce}(T, I_2, \tilde{I})]
    \end{aligned}
\end{equation}
Intuitively, by minimizing $\mathcal{L}_{cma}$, we encourage the visual features and linguistic features to be aligned well in the embedding space and in turn ease the feature fusion.

However, CMA loss \footnote{ALBEF \cite{li2021align} applies a special case of $\mathcal{L}_{cma}$ by setting $I_1=I_2$} ignores the self-supervision within each modality, thus failing to guarantee the desirable expressiveness of learned features. 
The reason is that i) text usually cannot fully describe the paired image.
For instance, although the text in Figure~\ref{fig:framework} (A) captures most of the salient objects in the image, it overlooks detailed features of each object, such as the cloth of the man.
Therefore, simply pulling embeddings of image-text pair together results in degraded representations (Figure~\ref{fig:framework} B); and
ii) image-text pairs used for pre-training are inherently noisy, which makes the problem in i) even worse.
To mitigate these limitations, we propose to further make use of intra-modal self-supervision by introducing Intra-Modal Contrastive (IMC) objective as follows.

\subsection{Intra-Modal Contrastive (IMC)}
Different from CMA, IMC attempts to learn the semantic difference between positive and negative samples within the same modality.
For the visual modality, we consider two random “views”  $(I_1, I_2)$ of the same image $I$ under random data augmentation as a positive pair.
Following \cite{he2020momentum,chen2020simple}, we maximize agreement between $(I_1, I_2)$ by using the contrastive loss $\mathcal{L}_{nce}(I_1, I_2, \tilde{I})$.
Similar to Equation \ref{eq:t2i}, we define $\text{sim}(I_1,I_2)=f_v(v_{cls})^T \hat{f}_v(\hat{v}_{cls})$.

For the text input, we follow \cite{gao2021simcse} to take a text and predict itself in a contrastive objective.
This is achieved by considering standard dropout as minimal data augmentation for the text, and applying independently sampled dropout masks for identical positive pairs, i.e., $T_+ = T$.
Different from \cite{gao2021simcse} that uses in-batch negatives, we use the same negative text queue $\tilde{T}$ in Equation \ref{eq:i2t} instead.
The contrastive objective can be described by $\mathcal{L}_{nce}(T, T_+, \tilde{T})$, where $\text{sim}(T,T_+)=f_t(t_{cls})^T \hat{f}_t(\hat{t}_{cls})$.
Overall, we minimize the following objective to guarantee reasonable intra-modal representation learning.
\begin{equation}
    \begin{aligned}
    \mathcal{L}_{imc}=\frac{1}{2}[\mathcal{L}_{nce}(T, T_+, \tilde{T})+\mathcal{L}_{nce}(I_1, I_2, \tilde{I})]
    \end{aligned}
\end{equation}
Specifically, our model is encouraged to learn representations that keep alignment between semantically-related positive pairs within a modality.
Most importantly, $\mathcal{L}_{imc}$ enforces the uniformity of the whole representation space of image and text such that the embeddings are uniformly distributed \cite{wang2020understanding}.
Therefore, CMA and IMC are designed to play a complementary role in the representation learning: i) CMA maps matched image-text pair close in the embedding space, and ii) IMC maximizes agreement between differently augmented views of the same data example.
Combining them together improves the quality of the learned representations (Figure~\ref{fig:framework} B) and can further facilitate joint multi-modal learning in the fusion encoder.

One limitation of IMC is that it simply performs the contrastive objective on [CLS] tokens of vision encoders and text encoders, where [CLS] tokens are assumed to represent the global information of the input.
In other words, IMC maximizes the global MI between differently augmented views.
However, the drawbacks of global MI maximization lie in that: i) it ignores the localized and structural information in the input \cite{hjelm2018learning,bachman2019learning}; ii) certain unrelated local regions may dominate the MI, resulting in the model that is biased to learning unrelated features.
For instance, noisy patches can represent a larger "quantity" of information than semantic-meaningful patches that occur repeatedly \cite{hjelm2018learning}.
To remedy this issue, we introduce local MI maximization into multi-modal representation learning as detailed below.

\subsection{Local MI Maximization (LMI)}
The goal of local MI maximization is to encourage high MI between the global representation and every local region (e.g., image patches and text tokens) of the input. 
Rather than considering the [CLS] token pair (e.g., $(v_{cls}, \hat{v}_{cls})$ of $(I_1, I_2)$) as a positive pair, we pair [CLS] token from one augmented version, with patch embeddings in the other independently augmented version of the input.
Without loss of generality, we take the vision input $(I_1, I_2)$ as an example.
Specifically, we consider $\{\hat{v}_i\}_{i=1}^M$ as positive examples of $v_{cls}$, while patch embeddings from other images in the same batch are used to build up negative examples.
Similarly, $\{\hat{t}_j\}_{j=1}^N$ are considered as positive examples of $t_{cls}$, while text tokens from other in-batch texts are negative examples.
We maximize the average MI between global and local regions by minimizing the following loss:
\begin{equation} \label{eq:lmi}
    \begin{aligned}
    \mathcal{L}_{lmi}=\frac{1}{2} \bigg[ \frac{1}{M} \sum_{i=1}^M \mathcal{L}_{nce}(I_1, I_2^i, \tilde{I_l}) + \frac{1}{N} \sum_{j=1}^N \mathcal{L}_{nce}(T, T_+^j, \tilde{T_l}) \bigg]
    \end{aligned}
\end{equation}
where $\text{sim}(I_1,I_2^i)=f_v(v_{cls})^T \hat{f}_v(\hat{v}_{i})$, $\text{sim}(T,T_+^j)=f_t(t_{cls})^T \hat{f}_t(\hat{t}_{j})$, $\tilde{I_l}$ and $\tilde{T_l}$ are in-batch negative image and text patch embeddings, respectively.
Therefore, minimizing $\mathcal{L}_{lmi}$ allows our model to encode the representations of data that are shared across all patches, rather than encoder representation from certain patches which dominate MI.
Another perspective is that local MI maximization encourages the model to predict local from the global representation, which forces the model to also capture fine-grained information and in turn to benefit joint representation learning.

\subsection{Image-Text Matching (ITM)}
To fuse vision and language representations, we adopt ITM which is widely used in previous VLP studies. 
Given an image-text pair, ITM predicts whether they are matched (positive examples) or not (negative examples), which can be regarded as a binary classification problem.
Following \cite{li2021align}, the fusion encoder takes $\{v_{cls}, v_1, ..., v_M\}$ and $\{{t}_{cls}, {t}_1, ..., {t}_N\}$ as input. 
We use [CLS] token of the fusion encoder as the joint representation of the input image-text pair, which is then fed into a fully-connected layer to predict the matching probability $\phi(I, T)$.
We assume that each image-text pair $(I, T)$ sampled from the pre-training datasets is a positive example (with label 1) and construct negative examples (with label 0) through batch-sampling \cite{li2021align}.
The ITM loss is defined as:
\begin{equation}
    \begin{aligned}
    \mathcal{L}_{itm}= \mathbb{E}_{p(I, T)} H(\phi(I,T), y^{(I,T)})
    \end{aligned}
\end{equation}
where $H(;)$ is the cross-entropy, $y^{(I,T)}$ denotes the label.

\subsection{Masked Language Modeling (MLM)}
We adopt MLM from BERT \cite{devlin2018bert}, which aims to predict the ground truth labels of masked text tokens $T^{msk}$.
Specifically, we randomly mask out text tokens with a probability of 15\%, and replace them with a special [MASK] token 80\% of the time, and 10\% with random words, and leave it unchanged for the remaining 10\% of the time \cite{devlin2018bert}.
Different from BERT, our MLM is conditioned on both surrounding text tokens of $T^{msk}$ and image representations.
The MLM loss is defined as:
\begin{equation}
    \begin{aligned}
    \mathcal{L}_{mlm}= \mathbb{E}_{p(I, T^{msk})} H(\Phi(I,T^{msk}), y^{T^{msk}})
    \end{aligned}
\end{equation}
where $\Phi(I,T^{msk})$ is the predicted probability of $T^{msk}$, and $y^{T^{msk}}$ is ground truth.

The overall training objective of our model is:
\begin{equation}
    \begin{aligned}
    \mathcal{L}= \mathcal{L}_{cma}+\mathcal{L}_{imc}+\mathcal{L}_{lmi}+\mathcal{L}_{itm}+\mathcal{L}_{mlm}
    \end{aligned}
\end{equation}
\section{Experiments}

\begin{table}
	\footnotesize
	\setlength\tabcolsep{8pt}
	\begin{center}
		\begin{tabularx}{.47\textwidth}{ c|ccccc @{} }
			\toprule
			& COCO & VG & SBU & CC & CC12M \\
			\midrule
			\# images & 113K & 100K & 859K & 2.92M & 10.97M \\
			\# text & 567K & 769K & 859K & 2.92M & 10.97M \\
			\bottomrule
		\end{tabularx}
	\end{center}
	\caption{Statistics of pre-training datasets.}
	\label{table:pretraining_data}
	\vspace{-0.2in}
\end{table}

\subsection{Pre-training Datasets}
Following previous experimental protocols \cite{chen2020uniter, li2021align}, we use COCO \cite{lin2014microsoft}, Visual Genome (VG) \cite{krishna2017visual}, Conceptual Captions (CC) \cite{sharma2018conceptual}, and SBU Captions \cite{ordonez2011im2text} as the pre-training dataset in our study, where a total of 4.0M unique images and 5.1M image-text pairs are covered.
We term this dataset as a 4M dataset in our study.
To prove that our method can be applied to large-scale datasets, we further use CC12M \cite{changpinyo2021conceptual}. 
Together with the 4M dataset, we, therefore, reach large-scale pre-training data with 14.97M unique images and 16M image-text pairs (Table~\ref{table:pretraining_data}).

\newcommand{\xmark}{\ding{55}}
\newcommand{\cmark}{\ding{51}}
\begin{table*}
	\footnotesize
	\setlength\tabcolsep{5pt}
	\begin{center}
		\begin{tabular}{l|c|cccccc|cccccc}
            \toprule 
                \multirow{3}{*}{Method} & \multirow{3}{*}{\#Images} & \multicolumn{6}{c}{MSCOCO (5K)} & \multicolumn{6}{c}{Flickr30K (1K)}\\
                
                & & \multicolumn{3}{c}{Text Retrieval} & \multicolumn{3}{c}{Image Retrieval} &
                \multicolumn{3}{c}{Text Retrieval} & \multicolumn{3}{c}{Image Retrieval} \\
            
                & & R@1 & R@5 & R@10 & R@1 & R@5 & R@10
                & R@1 & R@5 & R@10 & R@1 & R@5 & R@10 \\
                \midrule
                ImageBERT \cite{qi2020imagebert} & 6M & 44.0 & 71.2 & 80.4 & 32.3 & 59.0 & 70.2 & 70.7 & 90.2 & 94.0 & 54.3 & 79.6 & 87.5 \\
                
                UNITER \cite{chen2020uniter} & 4M & 64.1 & 87.7 & 93.3 & 48.8 & 76.7 & 85.8 & 80.7 & 95.7 & 98.0 & 66.2 & 88.4 & 92.9  \\

                ViLT \cite{kim2021vilt} & 4M & 56.5 & 82.6 & 89.6 &  40.4 & 70.0 & 81.1 & 73.2 & 93.6 & 96.5 &  55.0 & 82.5 & 89.8 \\
                
                CLIP \cite{radford2021learning} & 400M &  58.4 & 81.5 & 88.1 & 37.8 & 62.4 & 72.2 & 88.0 & 98.7 & 99.4 & 68.7 & 90.6 & 95.2 \\
                
                ALBEF \cite{li2021align} & 4M & 68.7 & 89.5 & 94.7 & 50.1 & 76.4 & 84.5 & 90.5 & 98.8 & \bf 99.7 & 76.8 & 93.7 & 96.7 \\
                
                \bf Ours & 4M & \bf 71.4 & \bf 90.8 & \bf 95.4 & \bf 53.5 & \bf 79.0 & \bf 87.1 & \bf 93.0 & \bf99.1 & 99.6 & \bf 79.6 & \bf 95.1 & \bf 97.4 \\
                
                \midrule
                ALIGN \cite{jia2021scaling}& 1.2B &  58.6 & 83.0 & 89.7 & 45.6 & 69.8 & 78.6 & 88.6 & 98.7 & 99.7 & 75.7 & 93.8 & 96.8 \\
                
                
                \bottomrule
            \end{tabular}
	\end{center}
	\caption{Performance comparison of zero-shot image-text retrieval on Flickr30K and COCO datasets. For completeness, we also provide the results of ALIGN \cite{li2021align} which uses 1.8B image-text pairs (1.2B unique images) for pre-training. For text-retrieval (TR) and image-retrieval (IR), we report the average of R@1, R@5 and R@10.}
	\label{table:zero_shot}
	\vspace{-0.1in}
\end{table*}


\begin{table*}
	\footnotesize
	\setlength\tabcolsep{5pt}
	\begin{center}
		\begin{tabular}{l|c|cccccc|cccccc}
            \toprule 
                \multirow{3}{*}{Method} &
                \multirow{3}{*}{\#Images} & \multicolumn{6}{c}{MSCOCO (5K)} & \multicolumn{6}{c}{Flickr30K (1K)}\\
                
                & & \multicolumn{3}{c}{Text Retrieval} & \multicolumn{3}{c}{Image Retrieval} &
                \multicolumn{3}{c}{Text Retrieval} & \multicolumn{3}{c}{Image Retrieval} \\
            
                & & R@1 & R@5 & R@10 & R@1 & R@5 & R@10
                & R@1 & R@5 & R@10 & R@1 & R@5 & R@10 \\
                \midrule
                ImageBERT \cite{qi2020imagebert} & 6M & 66.4 & 89.8 & 94.4 & 50.5 & 78.7 & 87.1 & 87.0 & 97.6 & 99.2 & 73.1 & 92.6 & 96.0 \\    
            
                UNITER \cite{chen2020uniter} & 4M & 65.7 & 88.6 & 93.8 & 52.9 & 79.9 & 88.0 & 87.3 & 98.0 & 99.2 & 75.6 & 94.1 & 96.8 \\
                
                VILLA \cite{gan2020large} & 4M & \xmark & \xmark & \xmark & \xmark & \xmark & \xmark & 87.9 & 97.5 & 98.8 & 76.3 & 94.2 & 96.8 \\
                
                OSCAR \cite{li2020oscar} & 4M & 70.0 & 91.1 & 95.5 &  54.0 & 80.8 & 88.5 & \xmark & \xmark & \xmark & \xmark & \xmark & \xmark \\
                
                ViLT \cite{kim2021vilt} & 4M & 61.5 & 86.3 & 92.7  & 42.7 & 72.9 & 83.1 & 83.5 & 96.7 & 98.6 &  64.4 & 88.7 & 93.8 \\
                
                UNIMO \cite{li2020unimo} & 4M & \xmark & \xmark & \xmark & \xmark & \xmark & \xmark &  89.7 & 98.4 & 99.1 & 74.7 & 93.47 & 96.1 \\
                
                SOHO \cite{huang2021seeing} & 200K & 66.4 & 88.2 & 93.8 & 50.6 & 78.0 & 86.7 & 86.5 & 98.1 & 99.3 &  72.5 & 92.7 & 96.1 \\
                
                ALBEF \cite{li2021align} & 4M &  73.1 & 91.4 & 96.0 & 56.8 & 81.5 & 89.2 & 94.3 & 99.4 & \bf99.8 & 82.8 & \bf 96.7 & 98.4 \\
                
                \bf Ours & 4M & \bf75.6 & \bf92.8 & \bf96.7 & \bf59.0 & \bf83.2 & \bf89.9 & \bf 94.9 & \bf99.5 & \bf99.8 & \bf84.0 & \bf96.7 & \bf98.5\\
                
                \midrule
                ALIGN \cite{jia2021scaling} & 1.2B & 77.0 & 93.5 & 96.9 & 59.9 & 83.3 & 89.8 & 95.3 & 99.8 & 100.0 & 84.9 & 97.4 & 98.6 \\
                

                \bottomrule
            \end{tabular}
	\end{center}
	\caption{Performance comparison of fine-tuned image-text retrieval on Flickr30K and COCO datasets. For completeness, we also provide the results of ALIGN \cite{li2021align} which uses 1.8B image-text pairs (1.2B unique images) for pre-training.}
	\label{table:fine_tune}
	\vspace{-0.1in}
\end{table*}


\begin{table}
	\footnotesize
	\setlength\tabcolsep{2.8pt}
	\begin{center}
		\begin{tabular}{l|c|cccccc}
            \toprule 
                \multirow{2}{*}{Method} &
                \multirow{2}{*}{\#Images} & \multicolumn{2}{c}{VQA} & \multicolumn{2}{c}{NLVR$^2$} &
                \multicolumn{2}{c}{SNLI-VE}\\
                
                & & test-dev & test-std & dev & test-P & val & test \\
                \midrule
                OSCAR \cite{li2020oscar} & 4M & 73.16 & 73.44 & 78.07 & 78.36 & \xmark & \xmark \\

                UNITER \cite{chen2020uniter} & 4M & 72.70 & 72.91 & 77.18 & 77.85 & 78.59 & 78.28  \\

                ViLT \cite{kim2021vilt} & 4M & 71.26 & \xmark & 75.7 & 76.13 & \xmark & \xmark \\
                
                UNIMO \cite{li2020unimo} & 4M & 73.29 & 74.02 & \xmark & \xmark & 80.0 & 79.1 \\
                
                
                VILLA \cite{gan2020large} & 4M & 73.59 & 73.67 & 78.39 & 79.30 & 79.47 & 79.03 \\
                
                ALBEF \cite{li2021align} & 4M & 74.54 & 74.70 & 80.24 & 80.50 & 80.14 & \bf80.30  \\
                
                \bf Ours & 4M & \bf 74.90 & \bf74.92 & \bf80.54 & \bf81.33 & \bf80.51 & 80.29 \\
                \midrule
                
                VinVL \cite{zhang2021vinvl} & 6M & 75.95 & 76.12 & 82.05 & 83.08 & \xmark & \xmark \\
                
                \bottomrule
            \end{tabular}
	\end{center}
	\caption{Performance comparison on vision+language tasks.}
	\label{table:vqa}
	\vspace{-0.1in}
\end{table}

\subsection{Downstream Tasks}
\paragraph{Image-Text Retrieval} includes two tasks: (1) image as query and text as targets (TR); (2) text as query and image as targets (IR). 
The pre-trained model is evaluated on Flickr30K \cite{plummer2015flickr30k} and COCO \cite{lin2014microsoft} by following both fine-tuning and zero-shot settings.
For the fine-tuning setting, the pre-trained model is fine-tuned on the training data and evaluated on the validation/test data.
For the zero-shot setting, the pre-trained model is directly evaluated on the test data.
In particular, for zero-shot retrieval on Flickr30K, we follow \cite{li2021align} to evaluate the model fine-tuned on COCO.

\vspace{-10pt}
\paragraph{Visual Question Answering (VQA) \cite{goyal2017making}} 
aims to predict the answer given an image and a question (in text format), which requires an understanding of vision, language, and commonsense knowledge to answer.
We consider this task as a generation problem by following the same setting in \cite{li2021align}.
Specifically, an answer decoder is fine-tuned to generate the answer from the 3,192 candidates.

\vspace{-10pt}
\paragraph{Visual Entailment (SNLI-VE) \cite{xie2019visual}}
predicts whether a given image semantically entails a given text, which is a three-classes classification problem.
Specifically, the class or relationship between any given image-text pair can be entailment, neutral, or contradictory.
Compared with VQA, this task requires fine-grained reasoning. 

\vspace{-10pt}
\paragraph{Visual Reasoning (NLVR$^2$) \cite{suhr2018corpus}}
determines whether a natural language caption is true about a pair of photographs.
We evaluate our model on NLVR$^2$ dataset which contains 107,292 examples of human-written English sentences paired with web photographs. 
Since this task takes a text and two images as input, we extend our model by following \cite{li2021align}.

\subsection{Implementation Details}
\label{implementation_details}
All of our experiments are performed on 8 NVIDIA A100 GPUs with PyTorch framework \cite{paszke2017automatic}.
Our vision encoder is implemented by ViT-B/16 with 12 layers and 85.8M parameters.
Both the text encoder and the fusion encoder are implemented by a 6-layer transformer. 
They are initialized by the first 6 layers and the last 6 layers of BERT\textsubscript{base} (123.7M parameters), respectively. 
We set $K=65,536$ and $m=0.995$.
For the pre-training stage, the model is trained for 30 epochs with a batch size of 512.
We use mini-batch AdamW optimizer \cite{loshchilov2017decoupled} with a weight decay of 0.02.
The learning rate is initialized as $1e-5$ and is warmed up to $1e-4$ after 2,000 training iterations. 
We then decrease it by the cosine decay strategy to $1e-5$.
For data augmentation, a 256$\times$256-pixel crop is taken from a randomly resized image and then undergoes random color jittering, random grayscale conversion, random Gaussian Blur, random horizontal flip, and RandAugment \cite{cubuk2020randaugment}.
During the fine-tuning stage, the image resolution is increased to 384$\times$384 and the positional encoding is interpolated according to the number of image patches. 

\subsection{Evaluation on Image-Text Retrieval}
To assess the generalization of the learned representations, the common practice is to perform the zero-shot transfer of the trained model to downstream tasks.
We evaluate our model by benchmarking the zero-shot image-text retrieval tasks on Flickr30K and COCO datasets by following the standard evaluation protocol.
As shown in Table~\ref{table:zero_shot}, our approach achieves the best performance while outperforming the existing state-of-the-art by a large margin.
Compared with ViLT \cite{kim2021vilt} which directly uses a transformer encoder to model the interaction between word and image patch embeddings, we improve +9.5\% (average) on COCO and +12.2\% (average) on Flickr30K, revealing the necessity of conducting cross-modal alignment before fusion.
ALBEF \cite{li2021align} is closely related to our work, which aligns image and text embeddings first, then uses a fusion encoder to learn joint representations.
Furthermore, ALBEF shares the same pre-training datasets with our method, thus making them comparable.
However, ALBEF ignores the intra-modal supervision, therefore the expressiveness of the learned features cannot be guaranteed.
Compared with ALBEF, our method brings +2.7\% TR/R@1 boost and +3.4\% IR/R@1 boost on MSCOCO (5K) dataset by explicitly leveraging the intra-modal information from both global and local perspectives. 
Details of intra-modal representation analysis are referred to the supplementary.
It is worth mentioning that our method demonstrates a significant improvement over ALIGN \cite{jia2021scaling}, i.e., a mean of 79.5\% vs 70.9\% on COCO and 94.0\% vs 92.2\% on Flickr30K.
Note, ALIGN is pre-trained on 1.8B image-text pairs which is approximately 360$\times$ times more image-text pairs than our model. 
This observation suggests that our method is more data-efficient which is mainly attributed to the consideration of intra-modal supervision. 
Overall, the representations learned by our method are more general and transferable than existing baselines.

For fine-tuned experiments, we set up new benchmark results as shown in Table~\ref{table:fine_tune}. 
On the medium-sized COCO dataset, we surpass ALBEF \cite{li2021align} by 2.5\% absolute TR/R@1 and 2.2\% absolute IR/R@1, revealing that our model can further benefit from fully-supervised training.
We also compete favorably against prior baselines on the small-sized Flickr30K dataset. 
The only exception is ALIGN \cite{jia2021scaling}, which outperforms our method by +0.48\% averaged (89.69\% vs 89.21\%) on COCO and Flickr30K, while at the expense of huge computational resources.
This is especially problematic for the scenario/researchers with limited budgets.
We believe that our method can also largely benefit from a much larger pre-training dataset, which is evidenced in section \ref{ablation}.

\subsection{VQA, VE, and NLVR$^2$}
Table~\ref{table:vqa} shows the performance comparison on VQA, VE, and NLVR$^2$ which requires image+text as inputs. 
In other words, to make succeed in these tasks, the model is supposed to have the capability in learning joint multi-modal embeddings.
Among five out of six criteria, we deliver state-of-the-art results, implying that explicitly considering cross-modal alignment and intra-modal supervision contribute to the feature fusion.
Note that VinVL \cite{zhang2021vinvl} outperforms our method, the main reason is that its pre-training corpus contains visual QA datasets, including GQA \cite{hudson2019gqa}, VQA \cite{goyal2017making}, and VG-QAs.

\begin{table}
	\footnotesize
	\setlength\tabcolsep{3.0pt}
	\begin{center}
		\begin{tabular}{l|cccc|cccc}
            \toprule 
                \multirow{3}{*}{Module} &
                \multicolumn{4}{c}{Zero-Shot} & \multicolumn{4}{c}{Fine-Tune}\\
                
                & \multicolumn{2}{c}{MSCOCO} & \multicolumn{2}{c}{Flickr30K} &
                \multicolumn{2}{c}{MSCOCO} &
                \multicolumn{2}{c}{Flickr30K}\\
                
                & TR & IR & TR & IR & TR & IR & TR & IR \\
                \midrule
                CMA+ITM+MLM & 68.7 & 50.1 & 90.5 & 76.8 & 73.1 & 56.8 & 94.3 & 82.8 \\
                
                +IMC (w/o aug) & 71.1 & 52.2 & 92.0 & 78.6 & 75.0 & 58.6 & 94.5 & 82.9 \\
                
                +IMC & 71.4 & 53.3 & 92.1 & 78.9 & 75.6 & 58.8 & 95.1 & 83.1 \\
                
                +IMC+LMI (\bf Ours) & 71.4 & 53.5 & 93.0 & 79.6 & 75.6 & 59.0 & 94.9 & 84.0  \\
                
                \bottomrule
            \end{tabular}
	\end{center}
	\caption{Ablation study of each component on image-text retrieval tasks. The R@1 is reported. For CMA+ITM+MLM, we use the results in ALBEF \cite{li2021align}.} 
	\label{table:ablation_module}
	\vspace{-0.1in}
\end{table}

\begin{table}
	\footnotesize
	\setlength\tabcolsep{2.8pt}
	\begin{center}
		\begin{tabular}{c|c|cccc|cccc}
            \toprule 
                \multirow{3}{*}{Pooling} &
                \multirow{3}{*}{Intermediate} &
                \multicolumn{4}{c}{Zero-Shot} & \multicolumn{4}{c}{Fine-Tune}\\
                
                & & \multicolumn{2}{c}{MSCOCO} & \multicolumn{2}{c}{Flickr30K} &
                \multicolumn{2}{c}{MSCOCO} &
                \multicolumn{2}{c}{Flickr30K}\\
                
                & & TR & IR & TR & IR & TR & IR & TR & IR \\
                \midrule
                & & 71.5 & 52.9 & 92.4 & 79.1 & 75.7 & 58.6 & 94.6 & 83.3 \\
                
                & \cmark & 71.4 & 52.9 & 91.5 & 77.9 & 75.7 & 58.6 & 94.4 & 82.3 \\
                
                \cmark & \cmark & 71.8 & 53.2 & 93.2 & 79.2 & 75.6 & 58.7 & 94.8 & 82.8 \\
                
                \cmark & & 71.4 & 53.5 & 93.0 & 79.6 & 75.6 & 59.0 & 94.9 & 84.0
                \\
                
                \bottomrule
            \end{tabular}
	\end{center}
	\caption{Ablation study of image patch pooling and intermediate local feature on image-text retrieval. R@1 is reported.}
	\label{table:ablation_pool}
	\vspace{-0.2in}
\end{table}

\subsection{Ablation Study}
\label{ablation}
To learn the effectiveness of the newly proposed modules (i.e., IMC and LMI) in improving the multi-modal representation learning, we perform ablation studies on image-text retrieval tasks shown in Table~\ref{table:ablation_module}.
Since ALBEF \cite{li2021align} is implemented by using loss function CMA+ITM+MLM, we thus use the results in ALBEF as the baseline.
We investigate two choices of IMC, i) IMC (w/o aug): only random crop, random horizontal flip, and RandomAugment are applied to the input image and set $I_1=I_2$ by following ALBEF; ii) IMC: $I_1$ and $I_2$ are two augmented views of the input image with stronger data augmentation as discussed in section \ref{implementation_details}.
Both strategies can improve the performance by a large margin, while stronger augmentation works better that is consistent with previous studies \cite{chen2020improved,chen2020simple}.
The performance is further improved by incorporating LMI, indicating the importance of localized and structural information in representation learning.

During the pre-training stage, each image is split into 256 patches with a size $16 \times 16$. 
To rule out the probability that small image patches may not contain enough information for local MI maximization, we apply global average pooling to the last-layer patch embeddings $\{\hat{v}_1, ..., \hat{v}_M\}$, resulting in $M=16$ patches for LMI in Equation \ref{eq:lmi}.
To maintain the spatial relationship among patches, we reshape $\{\hat{v}_1, ..., \hat{v}_M\}$ to the original 3D image space then apply the pooling operation.
Notably, different from \cite{hjelm2018learning} which uses feature maps from the intermediate layer as local information, we use patch embeddings pulled from the last layer.
We examine these two choices on image-text retrieval tasks as shown in Table~\ref{table:ablation_pool} and observe the importance of image patch pooling.
In addition, the performance of using last-layer patches is comparable to, if not better than, using patches from intermediate layers (i.e., 9th layer in $\hat{g}(\cdot)$ and 4th layer in $\hat{h}(\cdot)$).
We suspect that the difference between features learned by CNNs and vision transformers lead to this observation \cite{raghu2021vision}.

\begin{table}
	\footnotesize
	\setlength\tabcolsep{2.8pt}
	\begin{center}
		\begin{tabular}{l|cccc|cccc}
            \toprule 
                \multirow{3}{*}{Module} &
                \multicolumn{4}{c}{Zero-Shot} & \multicolumn{4}{c}{Fine-Tune}\\
                
                & \multicolumn{2}{c}{MSCOCO} & \multicolumn{2}{c}{Flickr30K} &
                \multicolumn{2}{c}{MSCOCO} &
                \multicolumn{2}{c}{Flickr30K}\\
                
                & TR & IR & TR & IR & TR & IR & TR & IR \\
                \midrule
                +IMC (w/o aug) (4M) & 71.1 & 52.2 & 92.0 & 78.6 & 75.0 & 58.6 & 94.5 & 82.9 \\
                
                +IMC (w/o aug) (14M) & 72.7 & 54.1 & 94.6 & 83.6 & 77.9 & 60.9 & 96.2 & 86.0 \\
                
                \bottomrule
            \end{tabular}
	\end{center}
	\caption{Ablation study of the size of pre-training datasets. R@1 is reported.}
	\label{table:ablation_14m}
	\vspace{-0.2in}
\end{table}

To study the impact of training on larger-scale datasets, we perform an ablation study on 14M datasets by using +IMC (w/o aug) as shown in Table~\ref{table:ablation_14m}.
We could clearly see that the larger scale dataset gave a significant boost in performance.
We hypothesis that our model has the potential for further improvement if pre-trained on further large datasets. 

We further investigate the importance of the momentum coefficient $m$ and observe that $m=0.5$ reaches the best performance (see supplementary).
This is different from MoCo \cite{he2020momentum} which claims that a reasonable momentum should be in 0.99$\sim$0.9999.
We leave this as our future work.

\section{Limitations}
The learned representations may tend to features present in the available data. If there are underrepresented groups, the model may be biased and perform worse on them.

\section{Conclusion}
In this paper, we propose a new vision-language pre-training framework named TCL (short for triple contrastive learning).
Different from previous studies that simply align image and text representations through a cross-modal contrastive loss, TCL further considers intra-modal supervision to guarantee that the learned representations are also meaningful within each modality, and in turn benefits cross-modal alignment and joint multi-modal embedding learning.
To incorporate the localized and structural information in representation learning, TCL further introduces the local MI which maximizes the mutual information between the global representation and the local information from image patches or text tokens.
Experimental results on widely used benchmarks show that TCL outperforms existing SOTA methods by a large margin.

\section{Acknowledgments}
This work was partially supported by US National Science Foundation IIS-1553687 and Cancer Prevention and Research Institute of Texas (CPRIT) award (RP190107).

{\small
\bibliographystyle{ieee_fullname}
\bibliography{egbib}
}

\end{document}